\documentclass[journal,twoside,web]{ieeecolor}
\usepackage{jsen}
\usepackage{cite}
\usepackage{amsmath,amssymb,amsfonts}
\usepackage{graphicx}
\usepackage{cuted}
\usepackage{textcomp}
\usepackage{tabularx,array}

\usepackage{xurl}
\usepackage[hidelinks]{hyperref}

\def\BibTeX{{\rm B\kern-.05em{\sc i\kern-.025em b}\kern-.08em
    T\kern-.1667em\lower.7ex\hbox{E}\kern-.125emX}}
\definecolor{abstractbg}{rgb}{0.89804,0.94510,0.83137}
\newcommand{\widetablecaption}[2]{%
  \refstepcounter{table}\label{#1}%
  {\color{subsectioncolor}\footnotesize TABLE~\thetable\par}%
  \vspace{1pt}%
  {\footnotesize\scshape #2\par}%
}
\newcommand{\codeyes}{\(\checkmark\)}
\newcommand{\codeno}{\(\times\)}

\begin{document}
\pagestyle{plain}
\title{4D Radar Perception Algorithms for Autonomous Driving: A Review}
\author{Xumin Wu, Jun Zhou, Jilin Mei, Chen Min, and Yu Hu
\thanks{Xumin Wu, Jun Zhou, Chen Min, and Yu Hu are with the Institute of Computing Technology, Chinese Academy of Sciences, Beijing 100190, China.}
\thanks{Jilin Mei is with Beijing Jingwei Hirain Technologies Co., Inc., Beijing, China.}
\thanks{Corresponding authors: Chen Min (mincheng@ict.ac.cn) and Yu Hu (huyu@ict.ac.cn).}}
\IEEEtitleabstractindextext{%
\fcolorbox{abstractbg}{abstractbg}{%
\begin{minipage}{\textwidth}%
\begin{abstract}
Research on 4D millimeter-wave radar perception algorithms has flourished in recent years, extending from signal processing and object detection to semantic segmentation, motion estimation, occupancy prediction, and dynamic scene reconstruction. This review organizes the field according to the evolution of perception tasks and algorithms. It first introduces radar fundamentals, data representations, and quality-enhancement methods, and then reviews object-level perception, motion and localization, local and dense spatial perception, and dynamic scene understanding. Across these directions, we compare radar-only learning, multimodal fusion, and cross-modal supervision and knowledge distillation. Particular attention is paid to how elevation, Doppler measurements, and radar physical priors are exploited across tasks. We further summarize the task coverage, input data, annotations, and evaluation protocols of existing datasets, clarifying the empirical support for different research directions. Finally, we discuss the common challenges and future directions of 4D radar perception for autonomous driving. This review provides a task-oriented perspective on the transition from sparse object perception to dynamic spatial understanding.
\end{abstract}

\begin{IEEEkeywords}
4D millimeter-wave radar, autonomous driving, Doppler velocity, dynamic scene understanding, multimodal perception, occupancy prediction, scene flow
\end{IEEEkeywords}
\end{minipage}}}

\maketitle
\thispagestyle{plain}

\color{nblue}
\section{Introduction}
\label{sec:introduction}
\color{black}
\IEEEPARstart{A}{utonomous} driving requires perception that remains reliable over long ranges, in dynamic traffic, and under adverse weather. Millimeter-wave radar is compact, cost-effective, robust under many adverse conditions, and directly measures radial velocity. Advances in multiple-input multiple-output (MIMO) arrays, imaging radar, and signal processing now enable elevation-resolved measurements, transforming radar from a target detector into a sensor that jointly observes 3D spatial structure and radial motion \cite{ref1,ref2}.

\begin{figure}[!t]
\centering
\includegraphics[width=0.68\columnwidth]{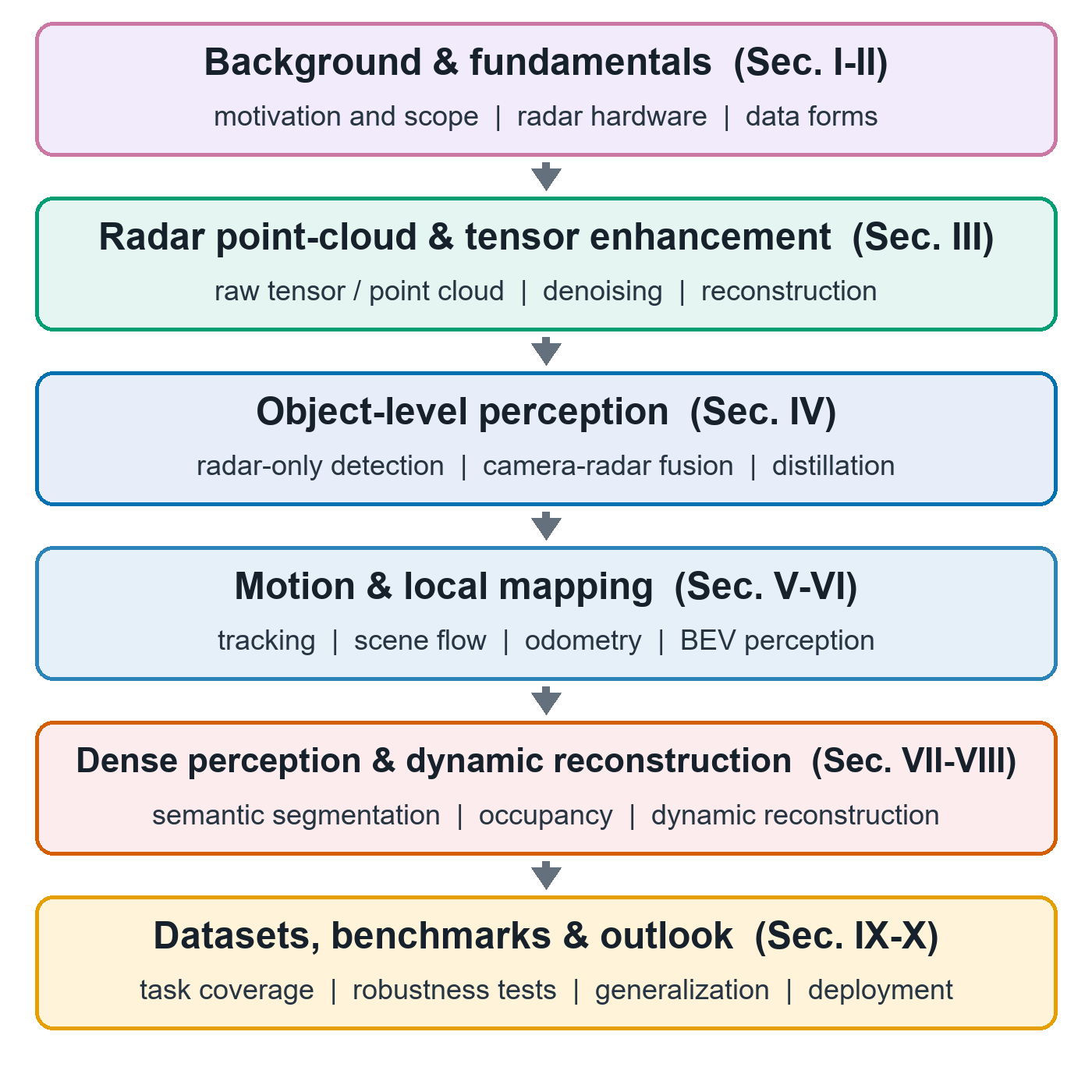}
\caption{Organization of this review. The paper introduces 4D radar fundamentals and data forms, then examines radar point-cloud and tensor enhancement, object-level perception, motion perception, local spatial perception, dense spatial perception, and dynamic dense scene understanding. It concludes with datasets, benchmark protocols, and future directions.}
\label{fig:review-organization}
\end{figure}

With this increase in spatial resolution, 4D radar research has expanded from vehicle detection to point-cloud enhancement, semantic segmentation, tracking, scene flow, ego-motion estimation, localization, occupancy prediction, and dynamic scene reconstruction. These tasks operate at different output levels, from individual returns and objects to local maps, occupied space, and temporally evolving scenes. A review of 4D radar perception should therefore explain how algorithms have evolved across these levels rather than treating detection as the sole downstream task. The organization of this review and the chronological progression of representative algorithms are shown in Figs.~\ref{fig:review-organization} and~\ref{fig:algorithm-timeline}, respectively.

Existing surveys cover FMCW principles and radar data representations \cite{ref1,ref52,ref55}, while broader reviews synthesize deep-learning radar perception and 4D radar applications \cite{ref51,ref53,ref56}. General automotive perception provides wider context \cite{ref54}, and radar--vision fusion has been reviewed separately \cite{ref57}. Prior 4D radar surveys focus mainly on fundamentals, representations, datasets, detection, and tracking \cite{ref1,ref51}; recent work on semantic segmentation, occupancy prediction, scene flow, and dynamic reconstruction has not yet been synthesized within a task-oriented framework. The cross-task use of elevation, Doppler, and radar physical priors, together with the empirical support provided by different datasets, also remains fragmented. This review addresses these gaps by organizing the progression from sparse object perception to dense and dynamic spatial understanding.

This review makes three contributions. First, it develops a task-oriented taxonomy that organizes 4D radar algorithms from point-cloud and tensor enhancement through object-level perception, motion and localization, local and dense spatial perception, and dynamic scene understanding; within these tasks, it compares radar-only learning, multimodal fusion, and cross-modal supervision, including knowledge distillation. Second, it analyzes how elevation, Doppler, RCS, radar measurement geometry, and motion consistency are used as physical priors or explicit constraints across perception tasks, and identifies where these cues remain underused. Third, it maps the task coverage, input data, annotations, and evaluation protocols of existing datasets, and identifies common challenges in data availability, physical fidelity, multimodal robustness, cross-platform generalization, and real-time deployment.

The literature search covered studies available through August 18, 2026. Google Scholar, IEEE Xplore, arXiv, and OpenAlex were searched, with backward reference tracing used to identify additional studies. We included automotive 4D imaging-radar research on data representation, object perception, motion estimation, occupancy prediction, and dynamic reconstruction, together with foundational measurement and dataset papers; non-automotive sensing was excluded. When a preprint and a published article reported the same study, the published version was retained.

\color{nblue}
\section{4D Radar Signals: Characteristics, Limitations, and Processing}
\label{sec:4d-radar-signals-characteristics-limitations-and-processing}
\color{black}
\subsection{Measurement Information and Signal Characteristics}
4D automotive radar is commonly described by four measurement dimensions: range, azimuth, elevation, and radial velocity. Its key capability is sufficiently resolved elevation combined with range, azimuth, and velocity, enabling returns to be localized in 3D space while radial motion is measured. Many sensors additionally report amplitude or radar cross section (RCS), whose value varies with target material, geometry, pose, and incidence angle \cite{ref1}. RCS has also been incorporated into Gaussian radar modeling and scan matching \cite{ref14}.

\subsection{Processing Chain and Algorithmic Inputs}
A minimal processing chain is ADC/I/Q samples $\rightarrow$ range and Doppler processing $\rightarrow$ angle estimation $\rightarrow$ radar tensors $\rightarrow$ CFAR or learned target extraction $\rightarrow$ point clouds or target lists. Range and Doppler fast Fourier transforms produce range-Doppler maps, while direction-of-arrival estimation or spatial Fourier transforms extend them to range-azimuth or range-azimuth-elevation tensors. CFAR, filtering, clustering, and postprocessing convert these dense responses into sparse point clouds. Downstream networks may further reorganize point clouds or tensors into bird's-eye-view (BEV), voxel, occupancy-grid, graph, or map representations \cite{ref55}.

Raw I/Q retains the most phase and weak-return information. These raw measurements contain the most complete information available to an algorithm, but they also depend strongly on waveform, antenna configuration, sampling strategy, and front-end calibration \cite{ref55}. Range-Doppler and range-azimuth-elevation tensors preserve structured spectral and spatial information and support early or intermediate learning. Point clouds are compact and compatible with LiDAR-inspired models, but CFAR thresholding removes sub-threshold responses before downstream learning \cite{ref55}. BEV and voxel features are task-oriented derived representations that facilitate fusion and planning. However, their construction can introduce quantization or completion errors.

Signal processing is mature in conventional automotive radar, whereas learning-based CFAR, angle super-resolution, and interference or clutter suppression remain rapidly developing directions for 4D imaging radar.

\subsection{Algorithm-Relevant Limitations}
Several limitations directly shape algorithm design. First, 4D radar returns remain sparse and nonuniform, and elevation accuracy is constrained by angular resolution, aperture, calibration, and sidelobes. Second, Doppler measures only radial velocity; lateral motion requires temporal association, geometry, or additional sensors. Third, multipath, clutter, sidelobes, interference, and target-dependent RCS can generate false, missing, or unstable returns. Radar is often more resilient than cameras or LiDAR in adverse weather \cite{ref2}. It is nevertheless not immune to precipitation and other noise sources, which can change point-cloud size and measurement variability \cite{ref79}.

\begin{strip}
\centering
\begin{minipage}{0.95\textwidth}
\centering
\includegraphics[width=0.84\textwidth]{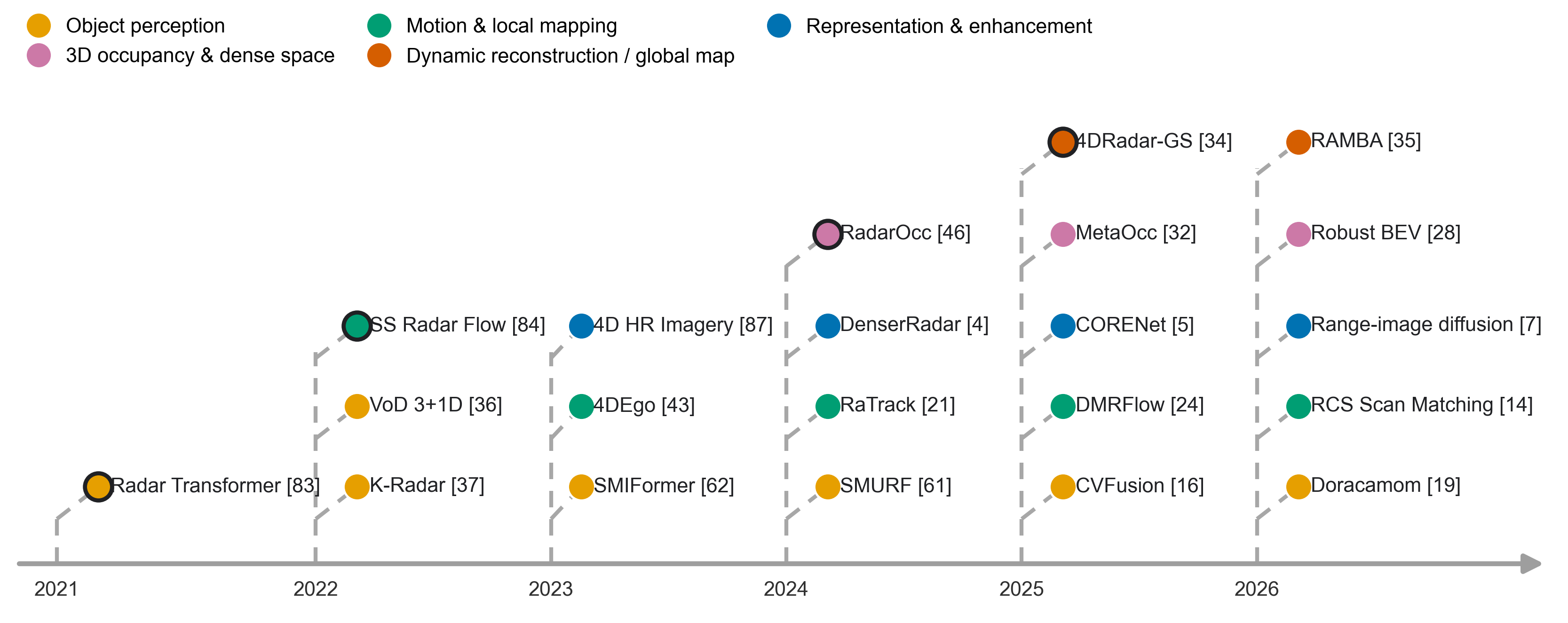}
\par
\refstepcounter{figure}
\footnotesize\sffamily
\textcolor{nblue}{Fig.~\thefigure.}\quad Timeline of representative automotive 4D imaging radar algorithms (2021--2026); black outlines mark verified category origins.
\label{fig:algorithm-timeline}
\end{minipage}
\end{strip}

Finally, radar inputs are not standardized across devices. Antenna layout, waveform, sampling rate, CFAR settings, coordinate conventions, and proprietary postprocessing alter point density and feature distributions \cite{ref15,ref55}. Algorithms trained on one sensor may therefore learn its processing pipeline rather than general radar structure. Studies should report the radar configuration and front-end processing and evaluate cross-device generalization in addition to downstream accuracy.

\begin{figure}[!h]
\centering
\includegraphics[width=0.62\columnwidth]{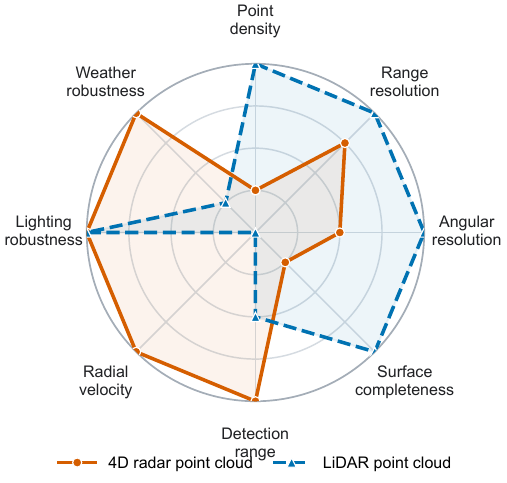}
\caption{Qualitative comparison of 4D-radar and LiDAR point clouds. Greater radial extent denotes a stronger characteristic. Sensor-level ratings are adapted from Table~I in \cite{ref51}, while point density and surface completeness summarize general point-cloud trends rather than a common benchmark.}
\label{fig:radar-lidar-point-cloud-comparison}
\end{figure}

\section{Radar Point-Cloud and Tensor Enhancement}
\label{sec:radar-point-cloud-and-tensor-enhancement}
\color{black}
Radar point-cloud and tensor enhancement is a moderately mature direction. As summarized in Fig.~\ref{fig:radar-lidar-point-cloud-comparison}, 4D radar point clouds are typically sparser and less geometrically complete than LiDAR point clouds, but they offer longer detection range, direct radial-velocity and RCS measurements, and stronger adverse-weather robustness \cite{ref1,ref51,ref55}. This complementarity motivates enhancement that improves geometric usability without discarding radar-specific motion and scattering evidence.

Current studies focus on denoising, densification, super-resolution, reconstruction, and synthetic radar-tensor generation. Their main learning mechanisms include cross-modal supervision, self-supervised representation learning, diffusion or adversarial generation, state-space modeling, and detection-guided reconstruction. Table~\ref{tab:enhancement-levels} organizes representative methods by the position at which they enhance the radar representation.

\begin{table}[!t]
\caption{Operational Taxonomy of 4D Radar Quality Enhancement}
\label{tab:enhancement-levels}
\centering
\setlength{\tabcolsep}{2.5pt}
\renewcommand{\arraystretch}{1.08}
\begin{tabularx}{0.99\columnwidth}{|>{\raggedright\arraybackslash}m{1.05cm}|>{\raggedright\arraybackslash}X|>{\raggedright\arraybackslash}m{2.60cm}|}
\hline
Level & Main operation & Representative methods \\
\hline
Input & Direct point-cloud or radar-tensor reconstruction and synthesis & R2LDM \cite{ref6}; range-image diffusion \cite{ref7}; L2RDaS \cite{ref9} \\
\hline
Feature & Denoising or representation enhancement & CORENet \cite{ref5}; Bootstrapping Radars \cite{ref42}; Radar-Mamba \cite{ref50} \\
\hline
Task & Enhancement optimized jointly with object detection & DenserRadar \cite{ref4}; Dual-View \cite{ref8}; SD4R \cite{ref29} \\
\hline
\end{tabularx}
\end{table}

\begin{figure}[!t]
\centering
\includegraphics[width=\columnwidth]{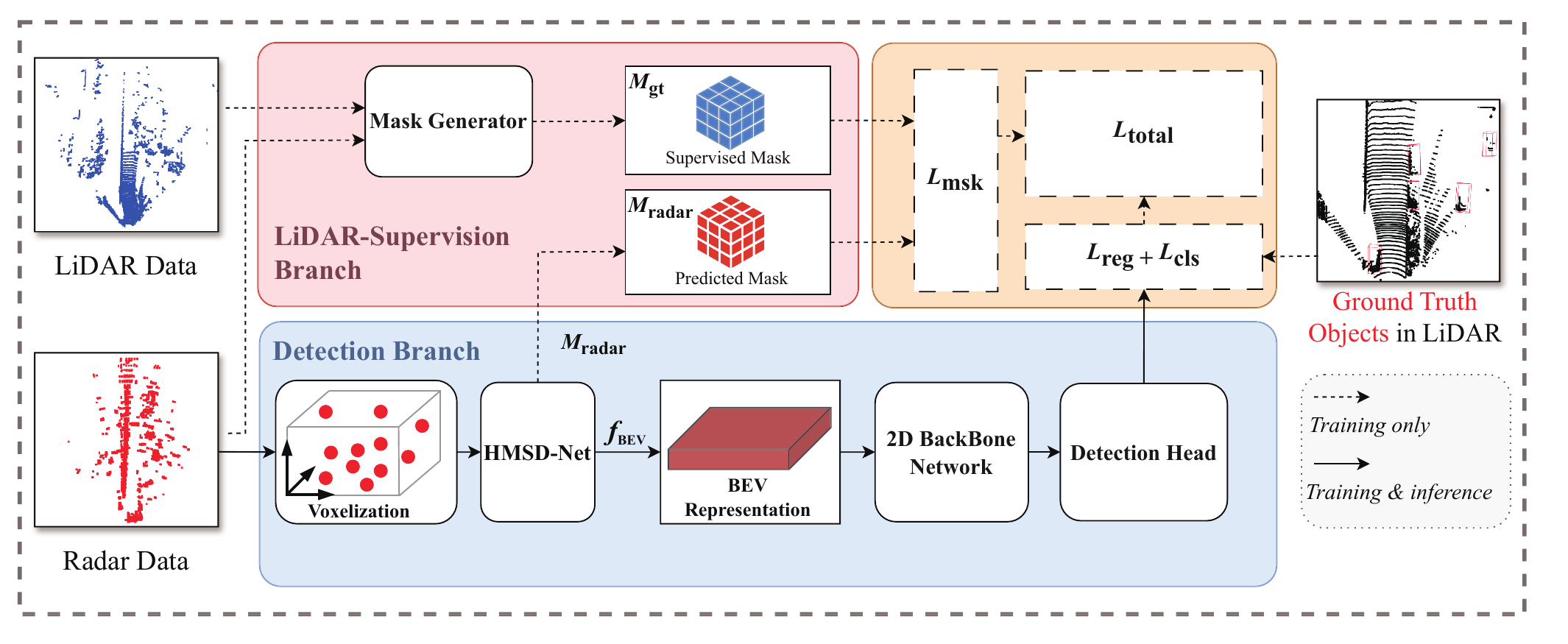}
\caption{Overall architecture of CORENet for LiDAR-supervised 4D radar denoising and object detection \cite{ref5}.}
\label{fig:corenet-framework}
\end{figure}

Cross-modal supervision uses a denser sensor during training to provide geometric targets while retaining radar-only inference. DenserRadar constructs dense 3D occupancy labels from temporally stitched LiDAR point clouds and trains a radar point-cloud detector to recover valid returns \cite{ref4}. The two-branch CORENet framework is shown in Fig.~\ref{fig:corenet-framework}. Its training-only LiDAR branch converts aligned LiDAR points into a voxel-level target mask through fixed-radius KD-tree matching. In the radar branch, voxelized returns pass through the hierarchical multi-scale denoising network, which combines point-topology and sparse-voxel features to predict which radar voxels should be retained. The predicted mask is supervised by the LiDAR-derived target mask, while the retained radar features are jointly optimized by the classification and regression losses of the detector. Inference therefore uses only the radar branch \cite{ref5}.

Self-supervised representation learning reduces dependence on point-level annotations. For each radar measurement, Bootstrapping Autonomous Driving Radars independently samples two transformations from an augmentation set to generate two radar heatmaps. The transformations include horizontal flipping, rotation and center cropping in radar coordinates, and Radar MIMO Mask. The two heatmaps derived from the same measurement form a positive pair, whereas heatmaps derived from different measurements in the same training mini-batch serve as negatives. Its intra-radar contrastive loss brings the positive radar views together and separates the negative samples. Its radar-to-vision loss further aligns each radar representation with the embedding of its synchronized camera frame and separates mismatched radar--image pairs. Radar MIMO Mask applies antenna dropout and random phase perturbations to MIMO channels to simulate incomplete apertures and Doppler phase distortion. The pretrained radar encoder is then adapted to downstream detection \cite{ref42}.

Generative methods explicitly reconstruct dense point clouds or synthesize radar tensors. R2LDM encodes paired radar and LiDAR point clouds as latent voxel features, conditions reverse diffusion on the radar features, and predicts nonempty voxels and within-voxel point offsets to produce a dense cloud. It reports a six- to ten-fold increase in point count together with gains in registration and detection \cite{ref6}. Range-image-driven diffusion instead projects radar points into a perspective range image and transfers priors from a pretrained image diffusion model before reconstructing the 3D cloud \cite{ref7}. L2RDaS uses a conditional generative adversarial network and an object-information supplement module to synthesize spatial radar tensors from LiDAR for dataset expansion \cite{ref9}.

Other methods enhance features or couple reconstruction directly to detection. Radar-Mamba aligns radar and LiDAR occupancies during training, uses state-space scans to capture local and global spatiotemporal features, and fuses Doppler and elevation features for denoising \cite{ref50}. Dual-View Radar Reconstruction predicts a BEV occupancy map and a perspective depth map from radar measurements, back-projects and fuses the two views under cross-view geometric consistency, and jointly encodes the original and reconstructed points for 3D detection \cite{ref8}.

Existing methods introduce concrete physical constraints primarily at the measurement and geometry levels. CORENet predicts a selection mask over measured radar voxels rather than synthesizing replacement coordinates, so retained points preserve their measured intensity and radial-velocity attributes \cite{ref5}. Bootstrapping Autonomous Driving Radars uses radar-coordinate transformations and MIMO-channel perturbations that preserve scene geometry while modeling antenna loss and Doppler-phase corruption \cite{ref42}. R2LDM reconstructs points on a shared spatial voxel grid, supervises nonempty-voxel masks, and bounds each predicted offset within its voxel; Dual-View Radar Reconstruction enforces geometric agreement between the back-projected BEV-occupancy and perspective-depth views \cite{ref6,ref8}. L2RDaS combines aligned real radar tensors, object-centered spatial anchors, and adversarial, feature-matching, and voxel-wise losses to constrain reflection distributions and spatial structure \cite{ref9}.

\color{nblue}
\section{Object-Level Perception}
\label{sec:object-level-perception}
\color{black}
Object-level perception is one of the most mature and extensively studied 4D radar directions. Research has progressed from early radar object classification \cite{ref83} to extensive 3D detection, with practical baselines established for both radar-only and fusion-based perception. Current work continues to improve recognition of vulnerable road users under sparse returns \cite{ref86}.

Current object-level studies follow three technical routes: radar-only learning, multimodal fusion, and cross-modal supervision. Radar-only learning uses radar during both training and inference; multimodal fusion uses cameras or LiDAR at both stages; cross-modal supervision uses auxiliary modalities during training but retains radar-only inference.

Radar-only detectors commonly encode point clouds as pillars, voxels, BEV maps, or multiple projected views. RadarNeXt combines a reparameterizable depthwise-convolution backbone with a multi-path deformable foreground-enhancement neck to suppress clutter and strengthen irregular foreground responses \cite{ref11}. 4DRadDet first forms potential object clusters and then uses their features as queries in cross-attention, allowing clustered returns to guide detection \cite{ref13}. SRFF fuses high- and low-level features with different sparsity to retain localization detail and semantic context for pedestrians and cyclists \cite{ref86}. MVFAN, SMURF, and SMIFormer further improve sparse-radar representation through multi-view or multi-representation interaction \cite{ref59,ref61,ref62}.

Multimodal fusion is currently the most active route in object-level perception, with camera--radar fusion accounting for the largest share of recent studies. According to sensor combination, these methods can be grouped into camera--radar, LiDAR--radar, and camera--LiDAR--radar fusion.

Camera--radar methods mainly use radar geometry to support image-view transformation and image semantics to compensate for sparse radar returns. GRC-Net extracts graph and pillar-based radar features and aligns them with image features in a common geometric space \cite{ref63}. MSSF alternates point--image interaction and multi-stage sampling, then uses semantic-guided voxel reweighting to reduce feature blurring \cite{ref60}. LXL predicts image-depth distributions and radar occupancy grids and uses the latter to guide image feature sampling \cite{ref64}. R4Det combines panoramic depth fusion, pose-independent temporal fusion, and instance-guided refinement to improve depth and small-object detection \cite{ref17}.

\begin{figure}[!t]
\centering
\includegraphics[width=\columnwidth]{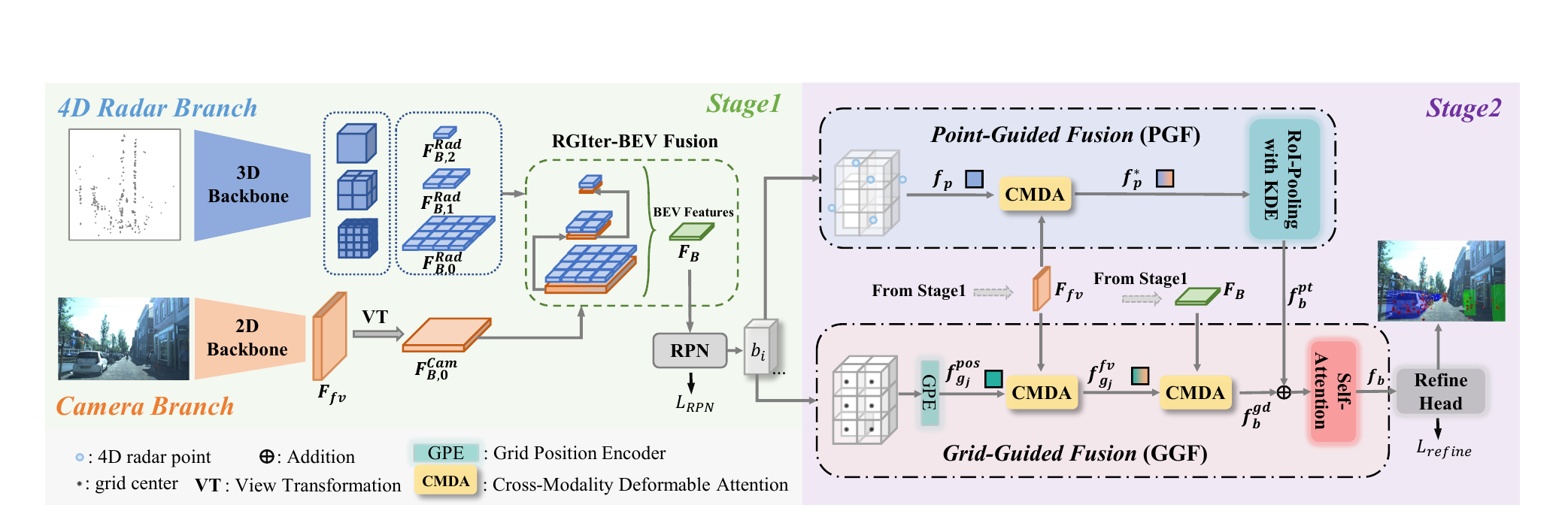}
\caption{Overall two-stage architecture of CVFusion for camera--4D-radar 3D object detection \cite{ref16}.}
\label{fig:cvfusion-framework}
\end{figure}

Other camera--radar methods focus on region selection and feature alignment. RPGFusion injects radar priors into multimodal feature selection \cite{ref18}; RCBEVDet++ performs dynamic alignment in BEV, whereas ZFusion explicitly models camera--radar feature interaction \cite{ref65,ref68}. Figure~\ref{fig:cvfusion-framework} shows CVFusion as a representative two-stage camera--radar detector. In the first stage, a radar-guided iterative BEV fusion module derives occupancy weights from radar features, uses them to progressively refine projected image BEV features, and generates high-recall proposals. In the second stage, point-guided and grid-guided branches aggregate radar-point, image-view, and BEV features within each proposal before final box refinement \cite{ref16}.

LiDAR--radar fusion directly combines complementary point-cloud measurements. $M^{2}$-Fusion uses interaction-based multimodal fusion to exchange intermediate radar and LiDAR features and center-based multi-scale fusion to extract object features at several resolutions \cite{ref58}. L4DR introduces multimodal encoding, foreground-aware denoising, and gated fusion to adapt to unequal sensor degradation under adverse weather \cite{ref66}. BSM-NET combines multi-bandwidth processing for point-density completion and denoising with multi-scale radar--LiDAR feature fusion \cite{ref67}.

Camera--LiDAR--radar fusion remains less developed in 4D radar research. LRVFNet separately encodes the three modalities, combines them through multi-scale attention, and applies a feature pyramid to multi-scale 2D object detection \cite{ref90}.

Cross-modal supervision transfers information from a denser modality without requiring that modality during deployment. HyperDet uses short-window radar accumulation, cross-sensor validation, and Doppler-guided motion compensation to construct enhanced radar inputs. LiDAR-guided pseudo-radar supervision enriches foreground geometry during training, while the measured radar background and radar-native attributes are retained; inference remains radar-only \cite{ref12}. SCKD uses a radar--LiDAR fusion teacher and a radar-only student. LiDAR-to-radar and fusion-to-radar feature distillation transfer teacher representations through separate adapters, while semi-supervised output distillation converts confident teacher predictions into pseudo-labels for labeled and unlabeled radar data \cite{ref20}.

Across these routes, radar physics is incorporated through measured attributes and geometric constraints. Elevation is retained through 3D coordinates, voxel features, and multi-view projections, whereas RCS and Doppler are used to distinguish returns and motion; MVFAN explicitly exploits RCS and Doppler, and 4DRadDet uses Doppler-supported clustering to construct object queries \cite{ref13,ref59}. HyperDet further applies Doppler-guided motion compensation and Doppler-consistent object relocation during temporal accumulation and augmentation \cite{ref12}. In fusion models, LXL uses radar occupancy to guide image-depth sampling, while CVFusion uses radar occupancy probabilities to constrain BEV features and proposals \cite{ref16,ref64}.

Table~\ref{tab:object-methods} compares representative methods from the three technical routes. All listed methods address 3D object detection, with SRFF focusing specifically on pedestrians and cyclists.

\begin{strip}
\centering
\widetablecaption{tab:object-methods}{Comparison of Representative 4D Radar Methods for 3D Object Detection}
\vspace{3pt}
\scriptsize
\setlength{\tabcolsep}{3pt}
\renewcommand{\arraystretch}{1.08}
\begin{tabularx}{\textwidth}{|>{\raggedright\arraybackslash}m{3.25cm}|>{\raggedright\arraybackslash}m{2.75cm}|>{\raggedright\arraybackslash}X|>{\centering\arraybackslash}m{1.55cm}|}
\hline
Method & Technical route & Core mechanism & Code available \\
\hline
SMIFormer \cite{ref62} (Sensors, 2023) & Radar-only learning & Multi-view interactive Transformer for spatial feature representation & \codeno \\
\hline
SRFF \cite{ref86} (Applied Sciences, 2024) & Radar-only learning & Fusion of high- and low-level features with different sparsity & \codeno \\
\hline
RadarNeXt \cite{ref11} (EURASIP JASP, 2025) & Radar-only learning & Reparameterized backbone and deformable foreground enhancement & \codeyes \\
\hline
4DRadDet \cite{ref13} (ICRA, 2025) & Radar-only learning & Candidate-object clustering and query-based detection & \codeno \\
\hline
LXL \cite{ref64} (IEEE T-IV, 2024) & Camera--radar fusion & Radar-occupancy-guided sampling of image-depth features & \codeyes \\
\hline
MSSF \cite{ref60} (IEEE T-ITS, 2025) & Camera--radar fusion & Point--image interaction and multi-stage sampling & \codeyes \\
\hline
CVFusion \cite{ref16} (ICCV, 2025) & Camera--radar fusion & Iterative BEV fusion and two-stage proposal refinement & \codeno \\
\hline
$M^{2}$-Fusion \cite{ref58} (IEEE TVT, 2023) & LiDAR--radar fusion & Intermediate feature interaction and multi-scale object fusion & \codeno \\
\hline
L4DR \cite{ref66} (AAAI, 2025) & LiDAR--radar fusion & Foreground-aware denoising and gated fusion & \codeyes \\
\hline
SCKD \cite{ref20} (AAAI, 2025) & Cross-modal supervision & Fusion-teacher and semi-supervised knowledge distillation & \codeno \\
\hline
\end{tabularx}
\vspace{1pt}
\begin{minipage}{\textwidth}
\footnotesize Note: \codeyes\ indicates that a method-specific official implementation or configuration was publicly available and verified on September 9, 2026; \codeno\ indicates that no such release was found.
\end{minipage}
\end{strip}

\color{nblue}
\section{Object Motion Perception}
\label{sec:object-motion-perception}
\color{black}
Motion perception comprises subdirections with unequal maturity. Multi-object and extended-target tracking form the most established branch. Moving-instance segmentation, ego-motion estimation, odometry, localization, and SLAM have reached an intermediate stage, whereas scene-flow estimation remains an early but active direction.

Object tracking is studied mainly through radar-only tracking and camera--radar fusion. Radar-only methods follow tracking-by-detection, extended-object filtering, or box-free moving-instance tracking. RaTrack combines motion segmentation, clustering, and motion estimation without category-specific boxes, whereas CenterRadarNet learns 3D detection and re-identification features from 4D radar tensors and uses the resulting embeddings for online association \cite{ref21,ref70}. Comparative studies evaluate tracking-by-detection frameworks, while extended-object filtering jointly estimates motion states and target contours from sparse radar measurements \cite{ref71,ref72}. In camera--radar tracking, CR3DT adds radar spatial and velocity features to a camera BEV detector, while disparity-domain depth recovery uses filtered radar points as metric anchors and performs cross-modal association at both detection and track levels \cite{ref48,ref82}.

Moving-instance segmentation and ego-motion estimation are increasingly addressed with radar-only temporal models. RadarMOSEVE uses spatial--temporal attention to jointly segment moving points and estimate ego velocity, while Radar Instance Transformer preserves full-resolution sparse features and assigns class-agnostic identities to moving instances \cite{ref45,ref73}. 4DEgo instead regresses ego velocity from high-resolution radar heatmaps through three-dimensional convolution and attention \cite{ref43}.

Odometry and SLAM rely more strongly on inertial or visual assistance. 4D iRIOM combines scan-wise ego-velocity estimation, scan-to-submap registration, inertial filtering, and loop closure in a radar--inertial SLAM system \cite{ref75}. DGRO and Go-RIO couple radar velocity with inertial measurements through graph-based preintegration and Gaussian-process continuous-time preintegration, respectively \cite{ref44,ref85}. Gaussian radar--inertial odometry models the scene with three-dimensional Gaussian distributions and evaluates multiple registration hypotheses \cite{ref22}. RDynaSLAM follows a different route: it projects radar-derived dynamic clusters into camera images and removes the corresponding visual keypoints before pose estimation \cite{ref23}.

\begin{figure}[!t]
\centering
\includegraphics[width=\columnwidth]{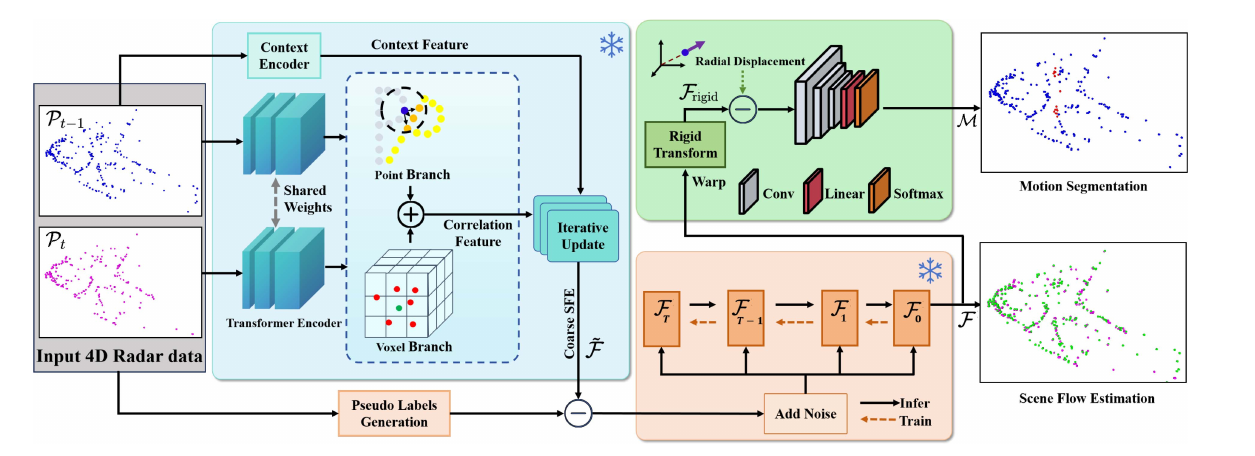}
\caption{Self-supervised scene-flow and motion-segmentation framework of RadarSFEMOS. Correlation features are constructed from adjacent 4D radar point clouds; coarse scene flow is first estimated and then refined by a diffusion model, while rigid-motion residuals are used to segment moving objects \cite{ref25}.}
\label{fig:radarsfemos-framework}
\end{figure}

Scene flow estimates point-wise three-dimensional motion and is studied through three main routes. Radar-only methods include self-supervised learning for sparse point clouds and decoupled correspondence and refinement in DMRFlow \cite{ref24,ref84}. RadarSFEMOS performs diffusion-based joint flow estimation and motion segmentation; its pipeline is illustrated in Fig.~\ref{fig:radarsfemos-framework} \cite{ref25}. RadarMP jointly generates radar points and estimates flow from consecutive low-level radar echoes, using radar-derived self-supervision rather than point-wise flow annotations \cite{ref27}. At inference time, RaLiFlow fuses radar and LiDAR with dynamic-aware bidirectional cross-modal interactions, whereas Hidden Gems uses colocated sensor cues only as cross-modal supervision for radar scene-flow training \cite{ref26,ref74}.

The physics-based priors and constraints used by these algorithms are derived from Doppler and RCS observations, rigid-motion consistency, and temporally coupled poses. RadarMOSEVE incorporates measured radial velocity into radar self- and cross-attention, while DGRO estimates ego velocity from Doppler, removes dynamic returns, and weights scan-to-submap registration by RCS \cite{ref44,ref45}. DMRFlow decouples positional and radial-velocity correspondence before refining static and dynamic flow separately; RadarSFEMOS derives motion-segmentation supervision from the discrepancy between robust rigid ego-motion and predicted scene flow; and RadarMP constructs spatial and motion consistency losses from Doppler shifts and echo intensity \cite{ref24,ref25,ref27}. For temporal pose estimation, 4D iRIOM combines robust single-scan velocity estimation with inertial filtering and loop closure, whereas Go-RIO uses continuous-time preintegration to accommodate asynchronous radar and IMU measurements \cite{ref75,ref85}.

Table~\ref{tab:motion-methods} compares representative motion-perception methods across tracking, motion segmentation, ego-motion estimation, odometry, SLAM, and scene flow.

\begin{strip}
\centering
\widetablecaption{tab:motion-methods}{Comparison of Representative 4D Radar Methods for Motion Perception}
\vspace{3pt}
\scriptsize
\setlength{\tabcolsep}{2.6pt}
\renewcommand{\arraystretch}{1.06}
\begin{tabularx}{\textwidth}{|>{\raggedright\arraybackslash}m{3.0cm}|>{\raggedright\arraybackslash}m{2.35cm}|>{\raggedright\arraybackslash}m{2.55cm}|>{\raggedright\arraybackslash}X|>{\centering\arraybackslash}m{1.45cm}|}
\hline
Method & Task & Technical route & Core mechanism & Code available \\
\hline
RaTrack \cite{ref21} (ICRA, 2024) & Moving-object tracking & Radar-only learning & Motion segmentation, clustering, and box-free tracking & \codeyes \\
\hline
CenterRadarNet \cite{ref70} (ICIP, 2024) & 3D detection and tracking & Radar-only learning & Joint detection and re-identification embeddings for association & \codeyes \\
\hline
CR3DT \cite{ref82} (IROS, 2024) & 3D detection and tracking & Camera--radar fusion & Camera BEV detection augmented by radar position and velocity & \codeno \\
\hline
RadarMOSEVE \cite{ref45} (AAAI, 2024) & Motion segmentation and ego velocity & Radar-only learning & Spatial--temporal Transformer for joint estimation & \codeyes \\
\hline
Radar Instance Transformer \cite{ref73} (IEEE T-RO, 2024) & Moving-instance segmentation & Radar-only learning & Full-resolution sparse features and class-agnostic instance identities & \codeno \\
\hline
4D iRIOM \cite{ref75} (IEEE RA-L, 2023) & Odometry and mapping & Radar--inertial fusion & Single-scan ego velocity, submap registration, filtering, and loop closure & \codeno \\
\hline
DGRO \cite{ref44} (Sensors, 2024) & Radar odometry & Radar--inertial fusion & Doppler--gyroscope preintegration and RCS-weighted registration & \codeno \\
\hline
RDynaSLAM \cite{ref23} (J. Intell. Robot. Syst., 2025) & Visual SLAM & Camera--radar fusion & Radar-derived dynamic masks for visual-keypoint removal & \codeno \\
\hline
DMRFlow \cite{ref24} (IEEE TCSVT, 2025) & Scene flow & Radar-only learning & Decoupled position--velocity matching and flow refinement & \codeno \\
\hline
RadarSFEMOS \cite{ref25} (IEEE RA-L, 2025) & Scene flow and motion segmentation & Radar-only learning & Rigid-motion self-supervision and diffusion-based refinement & \codeyes \\
\hline
RaLiFlow \cite{ref26} (AAAI, 2026) & Scene flow & LiDAR--radar fusion & Dynamic-aware bidirectional cross-modal interaction & \codeyes \\
\hline
Hidden Gems \cite{ref74} (CVPR, 2023) & Scene flow & Cross-modal supervision & Multisensor supervision for radar-only scene-flow learning & \codeyes \\
\hline
\end{tabularx}
\vspace{1pt}
\begin{minipage}{\textwidth}
\footnotesize Note: Code availability was verified against official repositories on September 9, 2026; symbol definitions are given in Table~II.
\end{minipage}
\end{strip}

\color{nblue}
\section{Local Spatial Geometry and Depth Completion}
\label{sec:local-spatial-perception}
\color{black}
Radar-guided depth completion and radar-based road-boundary perception remain early-stage 4D radar directions. Current work follows two main routes: camera--radar methods infer dense image-aligned metric depth from sparse radar ranges, whereas radar-only methods estimate local road-boundary geometry from temporal point clouds. Radar already measures range at observed returns; depth completion therefore fills unobserved image regions rather than re-estimating the measured radar range.

For depth completion, Semantic-Guided Depth Completion uses image semantics to diffuse sparse radar depth within category-specific regions. Its gradient-guided smooth-edge loss constrains consistency inside semantic regions and preserves depth discontinuities at their boundaries \cite{ref69}. SA-RCD uses RGB structure to define targeted support regions for sparse radar returns, enhances the radar depth representation, and fuses it with monocular-depth features through a multi-scale structure-guided network \cite{ref30}. In both methods, measured radar range establishes metric scale, while image structure supplies dense spatial support.

Radar-only local geometry is represented by 4DRadarRBD. It first removes unlikely returns using physical road-boundary constraints and then penalizes predicted points according to their distance from the annotated boundary. For temporal consistency, each radar point is augmented with its deviation from the motion-compensated boundary estimate of the previous frame before point-wise boundary segmentation \cite{ref49}.

\color{nblue}
\section{Dense Spatial Perception}
\label{sec:dense-spatial-perception}
\color{black}
Within dense spatial perception, point-wise semantic segmentation remains an early-stage direction, whereas grid- and voxel-based occupancy prediction has developed rapidly into a major 4D radar task. Point-wise segmentation assigns semantic labels only to measured radar returns, while occupancy prediction estimates free, occupied, and semantic states on a regular 3D grid, including locations without a return.

Existing point-wise segmentation studies are dominated by cross-modal knowledge distillation. RaSS introduces ZJUSSet with point-level radar and LiDAR labels and trains a radar-only student from a LiDAR segmentation teacher. Adaptive Doppler compensation aligns motion cues before segmentation \cite{ref31}. MKFusion likewise transfers knowledge from a multimodal teacher to improve radar point-cloud segmentation \cite{ref81}.

\begin{figure}[!t]
\centering
\includegraphics[width=\columnwidth]{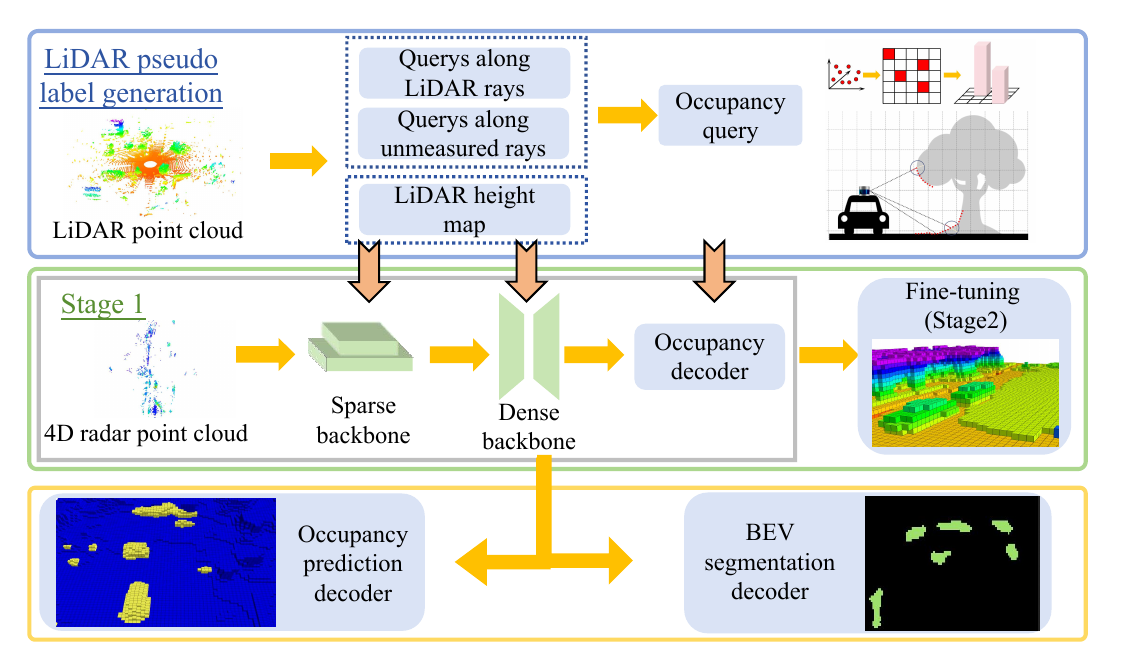}
\caption{Radar occupancy learning framework of 4D-ROLLS. LiDAR rays and height maps generate occupancy pseudo-labels; a sparse-to-dense backbone trains a radar-only occupancy network that jointly outputs occupancy prediction and BEV segmentation \cite{ref10}.}
\label{fig:rolls-occupancy}
\end{figure}

Occupancy methods differ mainly by sensor availability during training and inference. RadarOcc uses radar at both stages. It directly processes 4D radar tensors using Doppler-bin descriptors, sidelobe-aware spatial sparsification, and range-wise self-attention; spherical encoding preserves the measurement geometry before Cartesian voxel aggregation \cite{ref46}. 4D-ROLLS uses LiDAR only during training and remains radar-only at inference; Fig.~\ref{fig:rolls-occupancy} summarizes its supervision and occupancy network \cite{ref10}.

Camera--radar occupancy methods differ in how radar evidence enters the 3D representation. 4DRC-OCC fuses both modalities in voxel space, combining radar range and angle cues with image semantics \cite{ref33}. MetaOcc uses radar-height self-attention and hierarchical local--global fusion across modalities and time; its pseudo-label pipeline is a separate semi-supervised strategy for reducing annotation requirements \cite{ref32}. Doracamom combines radar geometry and image semantics to initialize voxel queries, then refines temporal representations in BEV and voxel spaces for joint detection and semantic occupancy \cite{ref19}. 4DR360 instead treats occupancy as a persistent scene state and updates it through state-guided BEV enhancement and Doppler-guided temporal fusion \cite{ref88}.

LiDAR-derived pseudo-labels in 4D-ROLLS and image semantics in camera--radar models are auxiliary spatial priors rather than radar physics. Measured Doppler, elevation, and spherical-coordinate structure instead provide radar-derived motion or geometric priors, while range and angle supply geometric anchors in voxel fusion. Encoding these quantities as features does not by itself constitute a hard physical constraint unless measurement consistency is explicitly enforced. Because the absence of a radar return does not imply free space, dense outputs should retain a distinction among directly observed, model-inferred, and unknown regions.

Table~\ref{tab:dense-spatial-methods} compares representative methods for point-wise semantic segmentation and dense occupancy prediction.

\begin{strip}
\centering
\widetablecaption{tab:dense-spatial-methods}{Comparison of Representative 4D Radar Methods for Dense Spatial Perception}
\vspace{3pt}
\scriptsize
\setlength{\tabcolsep}{2.8pt}
\renewcommand{\arraystretch}{1.08}
\begin{tabularx}{\textwidth}{|>{\raggedright\arraybackslash}m{3.0cm}|>{\raggedright\arraybackslash}m{2.45cm}|>{\raggedright\arraybackslash}m{2.55cm}|>{\raggedright\arraybackslash}X|>{\centering\arraybackslash}m{1.45cm}|}
\hline
Method & Task & Technical route & Core mechanism & Code available \\
\hline
RaSS \cite{ref31} (Sensors, 2025) & Point-wise semantic segmentation & Cross-modal supervision & Knowledge distillation from a LiDAR teacher to a radar student & \codeno \\
\hline
MKFusion \cite{ref81} (KBS, 2026) & Point-wise semantic segmentation & Cross-modal supervision & Semantic knowledge transfer from a multimodal teacher & \codeno \\
\hline
RadarOcc \cite{ref46} (NeurIPS, 2024) & 3D occupancy prediction & Radar-only learning & Doppler descriptors, sidelobe-aware sparsification, and spherical encoding & \codeyes \\
\hline
4D-ROLLS \cite{ref10} (IROS, 2025) & 3D occupancy and BEV segmentation & Cross-modal supervision & LiDAR-ray pseudo-labels for a radar-only occupancy network & \codeno \\
\hline
Doracamom \cite{ref19} (IEEE TCSVT, 2026) & Detection and semantic occupancy & Camera--radar fusion & Voxel-query initialization and temporal BEV--voxel fusion & \codeyes \\
\hline
\end{tabularx}
\vspace{1pt}
\begin{minipage}{\textwidth}
\footnotesize Note: Code availability was verified against official repositories on September 9, 2026; symbol definitions are given in Table~II.
\end{minipage}
\end{strip}

\color{nblue}
\section{Dynamic Dense Scene Understanding}
\label{sec:dynamic-dense-scene-understanding}
\begin{figure}[!t]
\centering
\includegraphics[width=0.68\columnwidth]{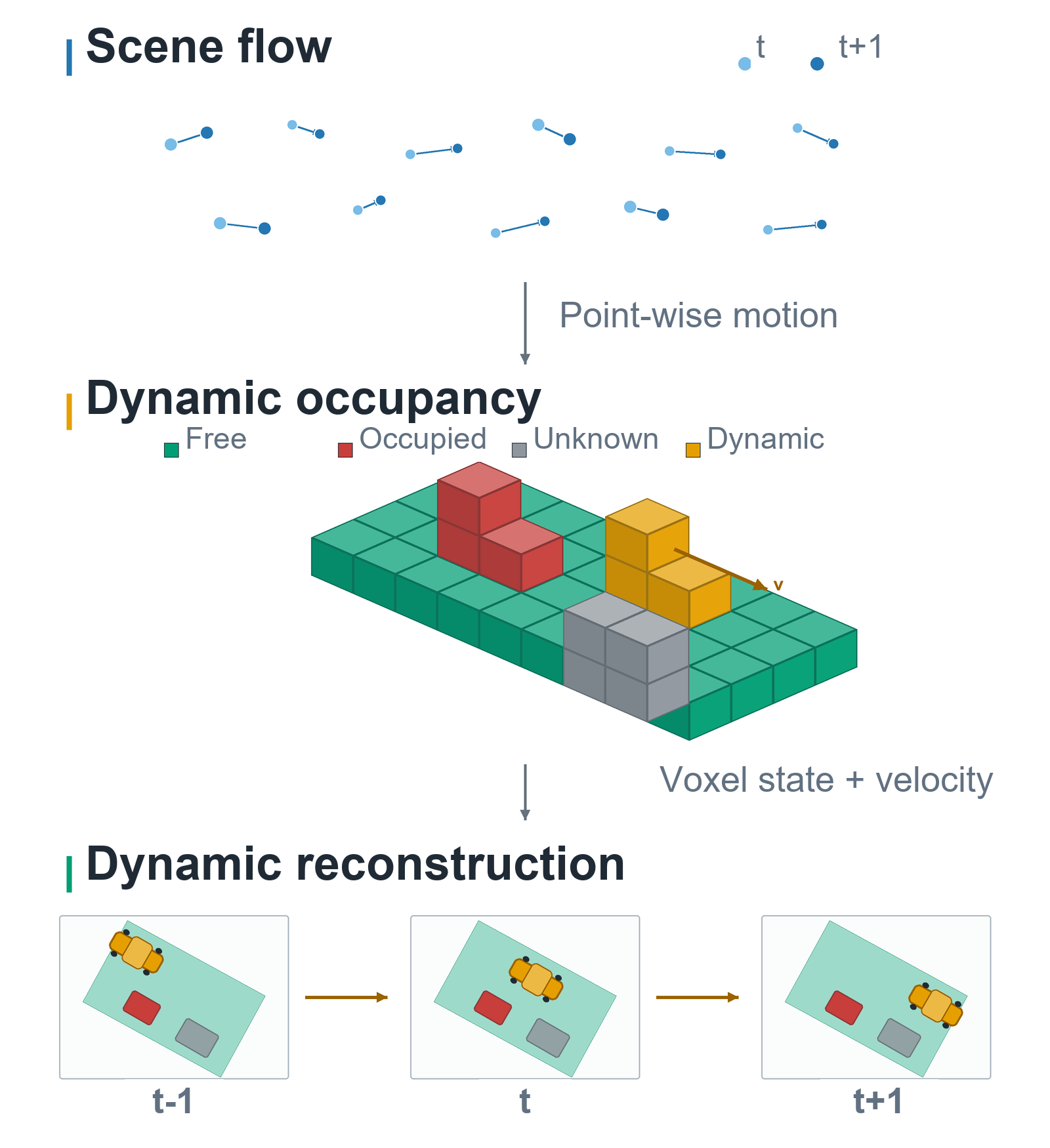}
\caption{Conceptual levels of dynamic dense scene understanding. Scene flow estimates sparse point-wise 3D motion, dynamic occupancy would associate voxel states with motion, and dynamic reconstruction maintains cross-frame trajectories and scene geometry.}
\label{fig:dynamic-scene-levels}
\end{figure}

\color{black}
Dynamic dense scene understanding is an exploratory direction that aims to represent both spatial structure and temporal evolution. Existing studies provide partial solutions at three levels---point-wise motion, temporally updated voxel states, and cross-frame scene reconstruction. However, a unified representation connecting these outputs has not yet emerged. Figure~\ref{fig:dynamic-scene-levels} summarizes these three conceptual levels.

At the point level, scene-flow methods recover the 3D motion of sparse radar returns between adjacent frames. These estimates provide motion cues for dense temporal reasoning, but they do not themselves encode free, occupied, dynamic, and unknown regions in a common spatial grid.

At the voxel level, existing 4D-radar occupancy methods mainly estimate geometric or semantic states, even when temporal fusion is used, and do not assign explicit motion to each voxel. Among them, 4DR360 is closest to persistent scene modeling because it maintains occupancy as a state across frames; its outputs, however, remain object detection and semantic occupancy rather than a velocity-aware dynamic occupancy field \cite{ref88}.

At the reconstruction level, 4DRadar-GS and RF4D represent two distinct technical routes. 4DRadar-GS adopts camera--radar fusion: radar spatial and velocity measurements guide dynamic-object segmentation, monocular-scale recovery, and Gaussian initialization, while a velocity-guided PointTrack model trained with scene-flow supervision maintains dynamic trajectories across frames \cite{ref34}. RF4D follows a radar-only route. Its time-conditioned neural field predicts scene-flow offsets to enforce temporal occupancy coherence and uses a radar-specific power-rendering formulation derived from the measurement process \cite{ref89}.

The two reconstruction routes also differ in output space. 4DRadar-GS produces image-oriented dynamic Gaussian representations, whereas RF4D synthesizes radar observations and occupancy in a continuous field; their evaluation targets are therefore not directly comparable. Cross-modal supervision and teacher--student distillation have not yet become established routes for this task.

This completes our review of the principal 4D radar perception directions and their representative methods.

\color{nblue}
\section{Datasets and Benchmark Protocols}
\label{sec:datasets-and-benchmark-evaluation}
\color{black}
Early 4D radar datasets concentrated on object-level perception. VoD and TJ4DRadSet established point-cloud benchmarks for 3D detection, while K-Radar added radar tensors and adverse-weather sequences. RaDelft subsequently provided lower-level radar data for automotive radar detection \cite{ref80}.

Task coverage later expanded beyond detection. ZJUSSet provides point-wise semantic labels, and the Radar-LiDAR Scene Flow Dataset repurposes synchronized VoD sequences with generated motion labels. DIDLM and NTU4DRadLM target localization, odometry, and mapping, whereas OmniHD-Scenes adds semantic occupancy to a full-surround radar-camera-LiDAR configuration. Dual Radar and V2X-Radar broaden multi-radar and cooperative perception, while DSERT-RoLL provides synchronized modalities, track identities, and ego odometry under diverse conditions. Table~\ref{tab:dataset-coverage} compares the official data forms and task annotations of the representative datasets retained for this analysis.

\begin{table*}[!t]
\centering
\caption{Official Data Forms and Task Coverage of Representative 4D Radar Datasets}
\label{tab:dataset-coverage}
\setlength{\tabcolsep}{2pt}
\footnotesize
\begin{tabularx}{0.98\textwidth}{|>{\raggedright\arraybackslash}m{1.85cm}|>{\raggedright\arraybackslash}X|*{3}{>{\centering\arraybackslash}m{2.05cm}|}>{\centering\arraybackslash}m{2.40cm}|}
\hline
Dataset & Radar form and synchronized modalities or labels & \shortstack{Object-level\\perception} & \shortstack{Motion,\\localization,\\and mapping} & \shortstack{Dense spatial\\perception} & \shortstack{Dynamic\\reconstruction} \\
\hline
VoD \cite{ref36} & 4D radar point clouds; LiDAR and stereo cameras & 3D detection & -- & -- & -- \\
\hline
K-Radar \cite{ref37} & 4D radar tensors; LiDAR, stereo cameras, and RTK-GPS & 3D detection & -- & -- & -- \\
\hline
TJ4DRadSet \cite{ref38} & 4D radar point clouds; synchronized camera & 3D detection & Object tracking & -- & -- \\
\hline
Dual Radar \cite{ref77} & Dual 4D radar point clouds; LiDAR and cameras & 3D detection & Object tracking & -- & -- \\
\hline
ZJUSSet \cite{ref31} & 4D radar and LiDAR point clouds; cameras and point-wise semantic labels & -- & -- & Point-wise semantic segmentation & -- \\
\hline
OmniHD-Scenes \cite{ref41} & Six radars, six cameras, and LiDAR; dense occupancy ground truth & 3D detection & -- & Semantic segmentation and occupancy & -- \\
\hline
DIDLM \cite{ref39} & Radar/LiDAR, RGB/infrared/depth, and IMU/GPS & -- & Localization, odometry, SLAM, mapping, and loop closure & -- & -- \\
\hline
NTU4DRadLM \cite{ref76} & Radar/LiDAR, RGB/thermal, and IMU/RTK & -- & Localization, odometry, mapping, and loop closure & -- & -- \\
\hline
V2X-Radar \cite{ref78} & Vehicle-side/roadside radars, LiDAR, and multi-view cameras & Cooperative, single-vehicle, and roadside 3D detection & -- & -- & -- \\
\hline
DSERT-RoLL \cite{ref40} & Radar, event/RGB/thermal cameras, dual LiDAR, and odometry ground truth & 2D/3D detection & Tracking and ego odometry & -- & -- \\
\hline
I/Q-1M$^{\dagger}$ \cite{ref3} & Complex range-Doppler-azimuth-elevation cubes; LiDAR and camera supervision & -- & Ego-motion estimation & BEV/3D occupancy and semantic segmentation & -- \\
\hline
\end{tabularx}
\vspace{1pt}
\begin{minipage}{\textwidth}
\footnotesize Note: $\dagger$ denotes a cross-domain raw-radar resource rather than a conventional road-vehicle dataset.
\end{minipage}
\end{table*}

Official task coverage does not exhaust later use. RaLiFlow repurposes synchronized VoD radar/LiDAR sequences by pairing adjacent frames and generating scene-flow labels during preprocessing. Its radar-denoising and label-generation procedure provides more reliable point-wise flow supervision for radar returns, especially those outside annotated object boundaries, and supports joint radar-LiDAR scene-flow learning \cite{ref26}. 4DRadar-GS uses synchronized radar-camera sequences from OmniHD-Scenes for dynamic reconstruction. It uses radar velocity and spatial information to segment dynamic objects and recover the monocular depth scale for Gaussian initialization, then jointly trains a velocity-guided point-tracking module under scene-flow supervision to maintain temporally consistent dynamic representations \cite{ref34}. These downstream protocols are therefore not assigned to the original VoD and OmniHD-Scenes rows in Table~\ref{tab:dataset-coverage}.

\begin{figure}[!t]
\centering
\includegraphics[width=\columnwidth]{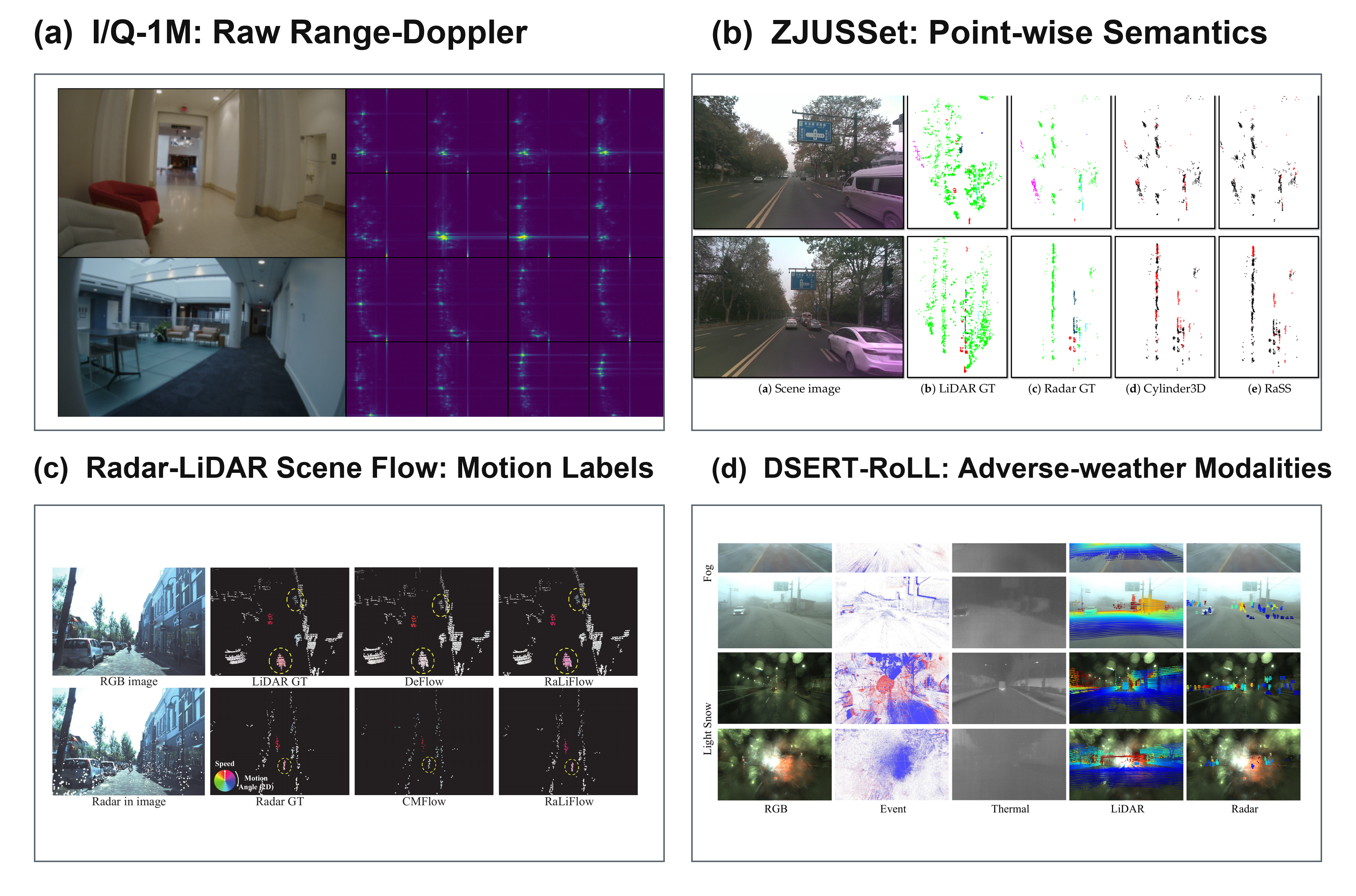}
\caption{Representative radar data forms and annotation strategies. Panels (a) and (b) show camera and range-Doppler frames from I/Q-1M and point-wise semantic labels from ZJUSSet, respectively \cite{ref3,ref31}. Panels (c) and (d) show VoD-derived radar-LiDAR scene-flow labels and synchronized adverse-condition modalities from DSERT-RoLL, respectively \cite{ref26,ref40}. Together, the panels contrast direct measurements with point-wise, generated-motion, and multimodal annotation strategies.}
\label{fig:dataset-data-forms}
\end{figure}

Existing benchmark protocols remain task-specific. Detection datasets report class-wise precision-recall and average precision, whereas semantic segmentation and occupancy use IoU-based metrics. Scene flow is evaluated by point-wise motion error, while localization and SLAM use trajectory and pose errors. These scores describe different outputs and cannot be compared as a single ranking of datasets or models.

Cross-dataset transfer is further complicated by front-end processing, point-cloud generation, sensor configuration, coordinate range, and annotation policy. Figure~\ref{fig:dataset-data-forms} visually illustrates differences in data representations and annotation strategies. As shown in Table~\ref{tab:dataset-coverage}, many road-driving datasets provide object annotations and postprocessed point clouds, whereas few combine measurement-level radar with dense spatial and motion labels. No single road-driving resource in the table jointly covers raw measurements, processed tensors, point clouds, object labels, dense occupancy, motion, full calibration metadata, and diverse radar hardware. This fragmentation makes it difficult to separate transferable radar representations from models fitted to a particular sensor and dataset protocol.

\color{nblue}
\section{Challenges and Future Directions}
\label{sec:challenges-and-future-directions}
\color{black}
First, measurement-level automotive radar data remain scarce because many road-driving datasets release processed point clouds rather than measurement-level I/Q data. Future datasets should jointly release raw I/Q, radar tensors, point clouds, BEV products, and synchronized auxiliary sensors together with the calibration, synchronization, waveform, array, CFAR, and point-cloud-generation parameters needed to reproduce front-end processing.

Second, cross-modal supervision should not be limited to LiDAR-like geometric completion. Although it can improve geometric completeness and dense prediction, treating unmatched radar returns indiscriminately as noise may suppress Doppler, RCS, multipath, and material-dependent cues. Future models should incorporate radar measurement equations, reflectivity models, Doppler consistency, multipath uncertainty, and sensor-noise statistics so that enhanced outputs remain compatible with the originating measurements.

Third, Doppler information remains underused. Existing studies exploit it in temporal detection, moving-object tracking, scene-flow estimation, and radar odometry. However, many methods still encode radial velocity only as an input feature, and current occupancy models generally do not explicitly estimate a velocity state for each voxel. Future research should make fuller use of Doppler information across detection, occupancy, motion, and mapping.

Fourth, multimodal fusion must progress from nominal accuracy gains to explicit robustness mechanisms. Real systems experience latency, extrinsic drift, sensor contamination, glare, precipitation, fog, and radar multipath. Radar-LiDAR calibration studies show that synchronization and extrinsic errors directly disrupt sparse cross-modal correspondence \cite{ref47}. Fog-aware radar-camera fusion has begun to address condition-dependent visual degradation \cite{ref28}. Future fusion models should adaptively weight modalities according to scene conditions and estimated reliability, with radar range and velocity contributing more strongly when visual reliability decreases or motion cues are critical.

Fifth, benchmark evaluation must become measurement-aware and task-specific. Generative enhancement, radar-data synthesis, and occupancy completion may infer spatial structure not directly supported by measured returns. Evaluation should therefore distinguish measured, inferred, free, occupied, and unknown regions and quantify velocity consistency, RCS preservation, uncertainty calibration, and planning impact. Current detection datasets commonly stratify results by class or range and, in K-Radar, by weather; future protocols should additionally stratify results by target velocity. Current scene-flow studies report point-wise flow errors; future motion evaluation should additionally report trajectory continuity and velocity consistency. SLAM and reconstruction should report pose error, map consistency, and dynamic-object handling.

Sixth, models must generalize across radar hardware, configurations, and vehicle platforms. Radar domain shift is strongly coupled to hardware because frequency band, antenna array, waveform, resolution, CFAR, and vendor postprocessing all alter the input. Sensor-parameter-conditioned encoding and multi-device pretraining are promising routes for exploiting raw and multi-sensor data. Evaluation should use cross-radar, cross-platform, and cross-city splits.

Seventh, real-time deployment and safety validation are essential. The engineering value of 4D radar lies in cost, environmental robustness, and manufacturability. Because diffusion, 3D voxel processing, large attention networks, and dynamic reconstruction add iterative or high-dimensional computation, their deployment cost should be reported explicitly. Future work should jointly report accuracy, latency, energy consumption, model size, calibration burden, failure detection, uncertainty, and planning impact on representative vehicle hardware.

From a system perspective, evaluation should extend beyond point density and aggregate mAP toward planning-relevant outputs. A useful 4D-radar interface should jointly represent objects; free, occupied, and unknown space; dynamic velocity; and uncertainty, while enforcing consistency among detection, occupancy, and motion estimates.

\color{nblue}
\section{Conclusion}
\label{sec:conclusion}
\color{black}
4D millimeter-wave radar perception is progressing from sparse object detection toward spatial and motion understanding. This review systematically organizes the field around radar point-cloud and tensor enhancement, object-level perception, motion and localization, local and dense spatial perception, dynamic scene reconstruction, datasets, and benchmark protocols. Object detection is relatively mature, occupancy prediction is developing rapidly, whereas point-wise semantic understanding and dynamic scene reconstruction remain exploratory. Radar-only learning, multimodal fusion, and cross-modal supervision are suited to different tasks and deployment conditions, yet radar-specific physical information---including elevation, Doppler, RCS, measurement geometry, and motion consistency---has not been fully or consistently exploited. Future research should expand measurement-level datasets, make fuller use of radar-specific physical information, and improve multimodal robustness, cross-device generalization, uncertainty evaluation, and efficient vehicle deployment.

\color{black}

\end{document}